\documentclass[runningheads, a4paper]{llncs}
\usepackage{amssymb, amsmath, bm, latexsym, comment}
\usepackage{multirow}
\usepackage{xcolor}
\usepackage{graphicx}
\usepackage{subfigure}
\usepackage{indentfirst}
\usepackage{setspace}
\usepackage{verbatim}
\usepackage{array}
\usepackage{cite}
\usepackage{booktabs}
\usepackage{hyperref}
\usepackage[capitalize]{cleveref}
\usepackage{arydshln}
\usepackage{multirow}

\newcolumntype{L}[1]{>{\raggedright\let\newline\\\arraybackslash\hspace{0pt}}m{#1}}
\newcolumntype{C}[1]{>{\centering\let\newline\\\arraybackslash\hspace{0pt}}m{#1}}
\newcolumntype{R}[1]{>{\raggedleft\let\newline\\\arraybackslash\hspace{0pt}}m{#1}}

\begin{document}

\title{MvMM-RegNet: A new image registration framework based on multivariate mixture model and neural network estimation}

\titlerunning{MvMM-RegNet}

\author{Xinzhe Luo \and
Xiahai Zhuang
\thanks{Corresponding author: zxh@fudan.edu.cn.}
}

\authorrunning{Xinzhe Luo et al.}

\tocauthor{}

\institute{School of Data Science, Fudan University, Shanghai, China}

\maketitle

\begin{abstract}
Current deep-learning-based registration algorithms often exploit intensity-based similarity measures as the loss function, where dense correspondence between a pair of moving and fixed images is optimized through backpropagation during training.
However, intensity-based metrics can be misleading when the assumption of intensity class correspondence is violated, especially in cross-modality or contrast-enhanced images.
Moreover, existing learning-based registration methods are predominantly applicable to pairwise registration and are rarely extended to groupwise registration or simultaneous registration with multiple images.
In this paper, we propose a new image registration framework based on multivariate mixture model (MvMM) and neural network estimation.
A generative model consolidating both appearance and anatomical information is established to derive a novel loss function capable of implementing groupwise registration.
We highlight the versatility of the proposed framework for various applications on multimodal cardiac images, including single-atlas-based segmentation (SAS) via pairwise registration and multi-atlas segmentation (MAS) unified by groupwise registration.
We evaluated performance on two publicly available datasets, i.e. MM-WHS-2017 and MS-CMRSeg-2019.
The results show that the proposed framework achieved an average Dice score of $0.871\pm  0.025$ for whole-heart segmentation on MR images and $0.783\pm 0.082$ for myocardium segmentation on LGE MR images\footnote{Code is available from \url{https://zmiclab.github.io/projects.html}.}.

\end{abstract}

\section{Introduction}\label{sec:introduction}
The purpose of image registration is to align images into a \emph{common} coordinate space, where further medical image analysis can be conducted, including image-guided intervention, image fusion for treatment decision, and atlas-based segmentation~\cite{journal/crp/Khalil2018}.
In the last few decades, intensity-based registration has received considerable scholarly attention.
Commonly used similarity measures comprise intensity difference and correlation-based methods for intra-modality registration, and information-theoretic metrics for inter-modality registration~\cite{journal/mp/Hajnal2001, journal/tmi/Mkel2002,
journal/tmi/Sotiras2013, journal/media/Viergever2016, journal/crp/Khalil2018}.

Recently, deep learning (DL) techniques have formulated registration as a parameterized mapping function, which not only made registration in one shot possible but achieved state-of-the-art accuracies~\cite{journal/media/Vos2018, journal/media/Hu2018,
journal/tmi/Balakrishnan2019, journal/media/Dalca2019}.
de Vos et al.~\cite{journal/media/Vos2018} computed dense correspondence between two images by optimizing normalized cross-correlation between intensity pairs.
While intensity-based similarity measures are widely used for intra-modality registration, there are circumstances when no robust metric, solely based on image appearance, can be applied.
Hu et al.~\cite{journal/media/Hu2018} therefore resorted to weak labels from corresponding anatomical structures and landmarks to predict voxel-level correspondence.
Balakrishnan et al.~\cite{journal/tmi/Balakrishnan2019} proposed leveraging both intensity- and segmentation-based metrics as loss functions for network optimization.
More recently, Dalca et al.~\cite{journal/media/Dalca2019} developed a probabilistic generative model and derived a framework that could incorporate both of the intensity images and anatomical surfaces.

Meanwhile, in the literature several studies have suggested coupling registration with segmentation, in which image registration and tissue classification are performed simultaneously within the same model~\cite{journal/ni/Ashburner2005, journal/ni/Pohl2006,
proceedings/miccai/Bhatia2007(2), journal/pami/Zhuang2019}.
However, the search for the optimal solution of these methods usually entails computationally expensive iterations and may suffer from problems of parameter tuning and local optimum.
A recent study attempted to leverage registration to perform Bayesian segmentation on brain MRI with an unsupervised deep learning framework~\cite{proceedings/miccai/Dalca2019}.
Nevertheless, it can still be difficult to apply unsupervised intensity-based approaches to inter-modality registration or to datasets with poor imaging quality and obscure intensity class correspondence.
Besides, previous DL-integrated registration methods have mainly focused on pairwise registration and are rarely extended to groupwise registration or simultaneous registration with multiple images.

In this paper, we consider the scenario in which multiple images from various modalities need to be co-registered simultaneously onto a \emph{common} coordinate space, which is set onto a reference subject or can be implicitly assumed during groupwise registration.
To this end, we propose a probabilistic image registration framework based on both multivariate mixture model (MvMM) and neural network estimation, referred to as MvMM-RegNet.
The model incorporates both types of information from the appearance and anatomy associated with each image subject, and explicitly models the correlation between them.
A neural network is then employed to estimate likelihood and achieve efficient optimization of registration parameters.
Besides, the framework provides posterior estimation for MAS on novel test images.

The main contribution of this work is four-fold.
First, we extend the conventional MvMM for image registration with multiple subjects.
Second, a DL-integrated groupwise registration framework is proposed, with a novel loss function derived from the probabilistic graphical model.
Third, by modelling the relationship between appearance and anatomical information, our model outperforms previous ones in terms of segmentation accuracy on cardiac medical images.
Finally, we investigate two applications of the proposed framework on cardiac image segmentation, i.e. SAS via pairwise registration and MAS unified by groupwise registration, and achieve state-of-the-art results on two publicly available datasets.

\section{Methods}\label{sec:methods}
Groupwise registration aims to align every subject in a population to a  \emph{common} coordinate space $\mathrm{\Omega}$
\cite{proceedings/miccai/Balci2007, proceedings/miccai/Bhatia2007(1)}, referred to as the \emph{common space}~\cite{journal/pami/Zhuang2019}.
Assume we have ${N_I}$ \emph{moving} subjects $\bm{I}=\{I_i\}_{i=1}^{N_I}$, of which each is defined on spatial domain $\mathrm{\Omega}_i$.
For each subject $I_i$, we can observe its appearance $A_i$ from medical imaging as well as labels of anatomical structures $S_i$ in various cases for image registration tasks.
Thus, we can formulate $I_i=(A_i, S_i)$ as a pair of appearance and anatomical observations for each subject.

\begin{figure}[t]
  \centering
  \subfigure[Groupwise registration]{
    \label{fig:groupwise_framework}
    \includegraphics[width=0.45\textwidth]{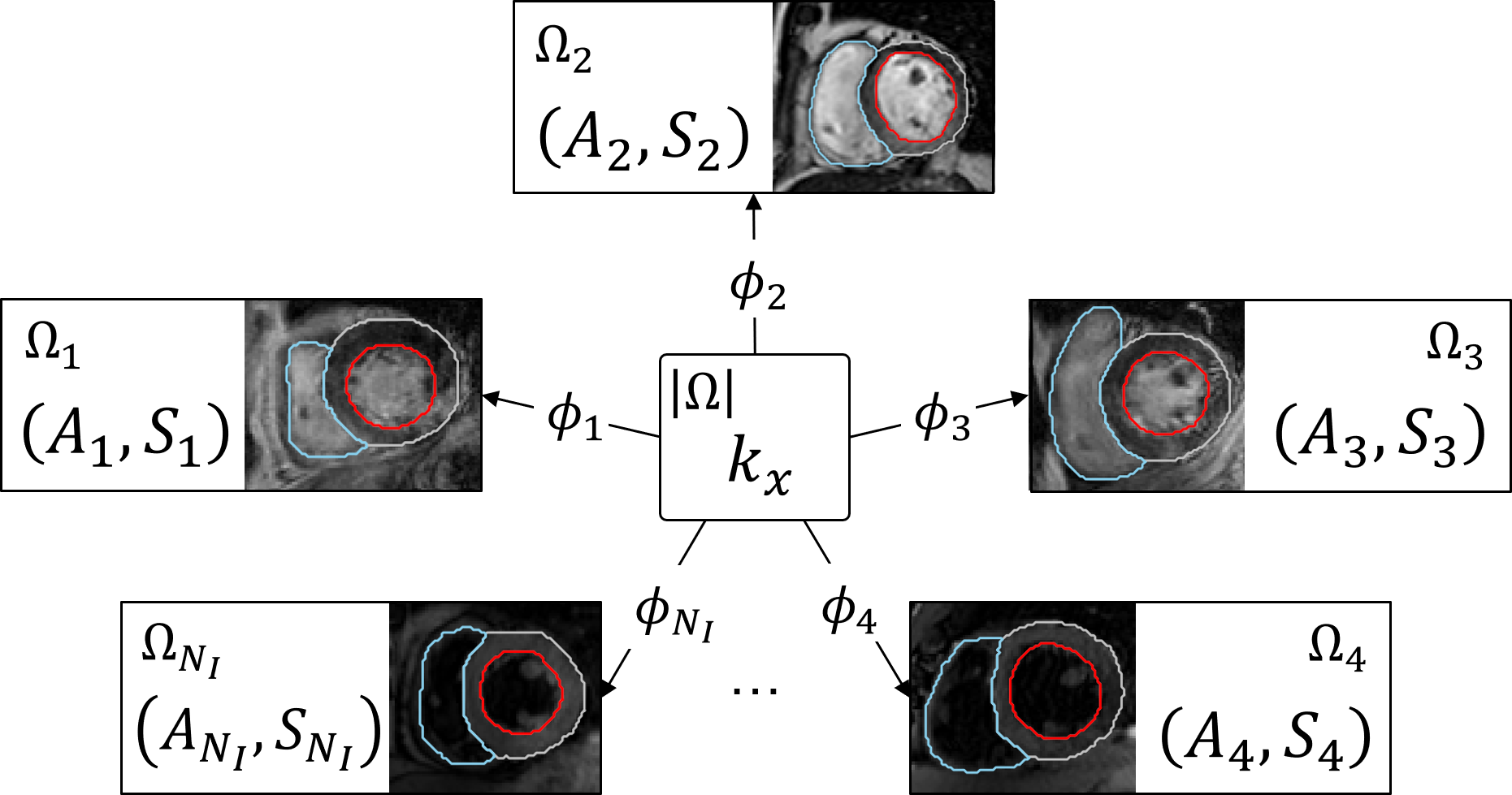}
  }
  \subfigure[Graphical model]{
    \label{fig:graphical_model}
    \includegraphics[width=0.45\textwidth]{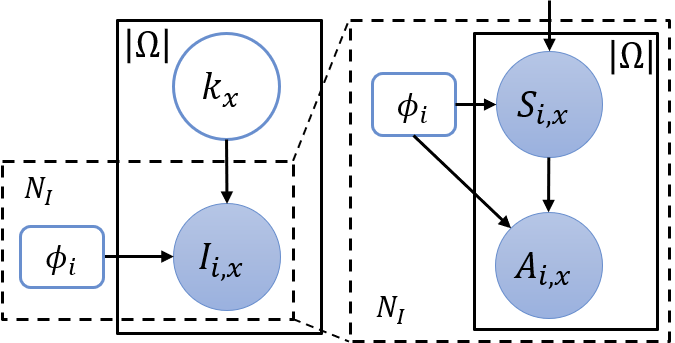}
  }
  \caption{(a) Groupwise registration framework, (b) Graphical representation of the proposed generative model, where random variables are in circles, deterministic parameters are in boxes, observed variables are shaded and plates indicate replication.}
\end{figure}

Associated with the \emph{moving} subjects is a set of spatial transforms $\bm{\phi}$ that map points from the common space to counterparts in each subject space:
\begin{equation}\label{eq:phi}
  \bm{\phi} = \{\phi_i: y_i=\phi_i(x),\ i=1, \dots, {N_I}\},
\end{equation}
where $x\in\mathrm{\Omega}$, $y_i\in\mathrm{\Omega}_i$.
The framework is demonstrated in \cref{fig:groupwise_framework}.

\subsection{Multivariate mixture model}\label{sec:mvmm}
The proposed method builds on a generative model of the appearance and anatomical information over a population of subjects.
The likelihood function is computed as a similarity measure to drive the groupwise registration process.

For spatial coordinates in the common space, an exemplar atlas can be determined \emph{a priori}, providing anatomical statistics of the population regardless of their corresponding appearances through medical imaging.
For notational convenience, we denote tissue types using label values $k_x$, where $k\in K$, $K$ is the set of labels, with its prior distribution defined as $\pi_{kx}=p(k_x)$.
Assuming \emph{independence} of each location, the likelihood can be written as $\mathcal{L}(\bm{\phi}|\bm{I})=\prod_{x\in\mathrm{\Omega}}p(\bm{I}_x|\bm{\phi})$.
Moreover, by summing over all states of the hidden variable $k_x$, we have
\begin{equation}
  \mathcal{L}(\bm{\phi}|\bm{I}) = \prod\limits_{x\in\mathrm{\Omega}}\sum\limits_{k\in K}\pi_{kx}\,p(\bm{I}_x|k_x,\bm{\phi}).
\end{equation}

Given the common-space anatomical structures, the multivariate mixture model  assumes \emph{conditional independence} of the moving subjects, namely
\begin{equation}
  p(\bm{I}_x|k_x, \bm{\phi})=\prod\limits_{i=1}^{N_I} p(I_{i, x}|k_x, \phi_i)
  =\prod\limits_{i=1}^{N_I} p(I_{i, y_i}|k_x),
\end{equation}
where $I_{i,y_i}$ denotes a patch of observations centred at $y_i=\phi_i(x)$.
Given anatomical structures of each subject, one can further assume its appearance is \emph{conditional independent} of the groupwise anatomy, i.e.
$p(A_{i,y_i}|S_{i,y_i},k_x) = p(A_{i,y_i}|S_{i,y_i})$.
Hence, we can further factorize the conditional probability into
\begin{equation}
  p(I_{i,y_i}|k_x) = p(A_{i,y_i}|S_{i,y_i})\,p(S_{i,y_i}|k_x).
\end{equation}
Accordingly, the log-likelihood is given by
\begin{equation}
    l(\bm{\phi}|\bm{I})=\sum\limits_{x\in\mathrm{\Omega}}\log\left\{
    \sum\limits_{k\in K}\pi_{kx}\prod\limits_{i=1}^{N_I}
    p(A_{i,y_i}|S_{i,y_i})\,p(S_{i,y_i}|k_x)
    \right\}.
\end{equation}
In practice, we optimize the negative log-likelihood as a dissimilarity measure to obtain the desired spatial transforms $\bm{\hat{\phi}}$.
The graphical representation of the proposed model is shown in \cref{fig:graphical_model}.

\subsection{The conditional parameterization}\label{sec:parameterization}
In this section, we specify in detail the conditional probability distributions (CPDs) for a joint distribution that factorizes according to the Bayesian network structure represented in \cref{fig:graphical_model}.

\subsubsection{Spatial prior.}
One way to define the common-space spatial prior is to average over a cohort of subjects~\cite{journal/ni/Ashburner2005}, and the resulting probabilistic atlas is used as a reference.
To avoid bias from a fixed reference and consider the population as a whole, we simply adopt a flat prior over the common space, i.e.\ $\pi_{kx}=c_k$, $\forall x\in \mathrm{\Omega}$ satisfying $\sum_{k\in K} c_k =1$, where $c_k$ is the weight to balance each tissue class.

\subsubsection{Label consistency.}
Spatial alignment of a group of subjects can be measured by their label consistency, defined as the joint distribution of the anatomical information $p(\bm{S}_{i,y_i}, k_x)$, where $\bm{S}_{i,y_i}=\{S_{i,y_i}\}_{i=1}^{N_I}$.
Each CPD $p(S_{i,y_i}|k_x)$ gives the likelihood of the anatomical structure around a subject location being labelled as $k_x$, conditioned on the transform that maps from the common space to each subject space.
We model it efficiently by a local Gaussian weighting:
\begin{equation}
  p(S_{i,y_i}|k_x) = \sum\limits_{z\in\mathcal{N}_{y_i}}w_z\cdot \delta(S_i(z)=k_x),
\end{equation}
where $\delta(\cdot)$ is the Kronecker delta function,  $\mathcal{N}_{y_i}$ defines a neighbourhood around $y_i$ of radius $r_s$ and $w_z$ specifies the weight for each voxel within the neighbourhood.
This formulation is equivalent to applying Gaussian filtering using an isotropic standard deviation $\sigma_s$ to the segmentation mask~\cite{journal/media/Hu2018}, where we set $r_s=3\,\sigma_s$.

\subsubsection{Appearance model.}\label{sec:appearance}
\begin{figure}[t]
  \centering
  \includegraphics[width=\textwidth]{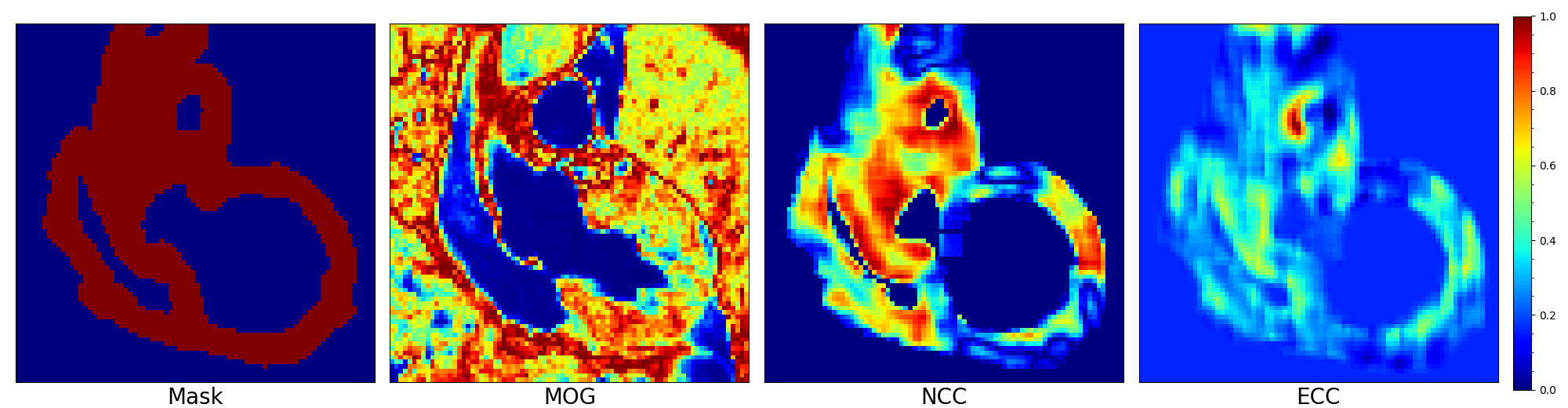}
  \caption{Visualization of different appearance models computed from a coronal view of a whole heart MR image subject at background areas, where "Mask", "MOG", "NCC" and "ECC" denote appearance model using ROI mask, mixture of Gaussians, normalized cross correlation and entropy cross correlation, respectively. For comparison, values are normalized to intervals between 0 and 1.}
  \label{fig:weight_jet}
\end{figure}

Finally, we seek to specify the term $p(A_{i,y_i}|S_{i,y_i})$.
A common approach adopted by many tissue classification algorithms
\cite{journal/tmi/Leemput1999, journal/ni/Ashburner2005,
journal/media/Iglesias2013, journal/pami/Zhuang2019,
proceedings/miccai/Dalca2019} is to model this CPD as a mixture of Gaussians (MOG), where intensities of the same tissue type should be clustered and voxel locations are assumed independent.
Nevertheless, we hypothesize that using such an appearance model can mislead the image registration when the assumption of intensity class correspondence is violated, due to poor imaging quality, particularly in cross-modality or contrast enhanced images~\cite{journal/tmi/Pluim2003}.
Let $\nabla(\cdot)$ and $\|\cdot\|$ be the voxel-wise gradient and Euclidean-norm operators, respectively.
A vanilla means is to use a mask around the ROI boundaries:
\begin{equation}
  p(A_{i,y_i}|S_{i,y_i})=\left\{
  \begin{aligned}
    &\ 1\quad \text{if}\ \exists\ z\in\mathcal{N}_{y_i}\ \text{s.t.}\ \nabla \|S_i(z)\|>0 \\
    &\ 0\quad \text{otherwise},
  \end{aligned}
  \right.
\end{equation}
which ignores the appearance information.
However, we argue that a reasonable CPD design should reflect fidelities of medical imaging and serve as a voxel-wise weighting factor for likelihood estimation.
Thus, we formalize a CPD that
1) is defined with individual subjects,
2) is zero on voxels distant to the ROIs,
3) has increasing values at regions where appearance and anatomy have consistent rate of change.
Therefore, we speculate that voxels with concordant gradient norms between appearance and anatomy are more contributory to determining the spatial correspondence.
Based on these principles, one can estimate the CPD as a Gibbs distribution computed from an energy function or negative similarity measure between gradient-norm maps of appearance and anatomy, i.e.
\begin{equation}
  p(A_{i,y_i}|S_{i,y_i}) = \frac{1}{Z} \exp\left\{-E(\|\nabla A_{i,y_i}\|, \|\nabla S_{i,y_i}\|)\right\},
\end{equation}
where $Z$ is the normalization factor and $E(\cdot)$ can be the negative normalized cross-correlation (NCC)~\cite{journal/media/Avants2008} or negative entropy correlation coefficient (ECC)~\cite{journal/tmi/Maes1997}. \cref{fig:weight_jet} visualises the different appearance models.

\subsection{Neural network estimation}\label{sec:network}
We formulate a neural network $g_{\bm{\theta}}(\cdot)$ parameterized by $\bm{\theta}$ that takes as input a group of $N_I$ images to predict the deformation fields, based on a 3D UNet-style architecture designed for image registration~\cite{journal/media/Hu2018}.
To discourage non-smooth displacement, we resort to bending energy as a deformation regularization term and incorporate it into the loss function~\cite{journal/media/Vos2018, journal/media/Hu2018}.
Hence, the final loss function for network optimization becomes
\begin{equation}\label{eq:loss}
  Loss(\bm{\theta};\bm{I}) = -l(\bm{\phi}_{\bm{\theta}}|\bm{I}) + \lambda\cdot R(\bm{\phi}_{\bm{\theta}}),
\end{equation}
where $R(\cdot)$ denotes the deformation regularization term and $\lambda$ is a regularization coefficient.

\subsection{Applications}\label{sec:applications}
In this section, we present two applications from the proposed MvMM-RegNet framework, which are validated in our experiments.
\subsubsection{Pairwise MvMM-RegNet for SAS.}
Pairwise registration can be considered as a specialization of groupwise registration where the number of subjects equals two and one of the spatial coordinate transforms is the identity mapping.
We will demonstrate the registration capacity of our model by performing pairwise registration on a real clinical dataset, referred to as pMvMM-RegNet.

\subsubsection{Groupwise MvMM-RegNet for MAS.}
During multi-atlas segmentation (MAS), multiple expert-annotated images with segmented labels, called atlases, are co-registered to a target space, where the warped atlas labels are combined by label fusion~\cite{journal/media/Iglesias2014}.
Delightfully, our model provides a unified framework for this procedure through groupwise registration, denoted as gMvMM-RegNet.
By setting the \emph{common space} onto the target as the \emph{reference space}, we can derive the following segmentation formula:
\begin{equation}
  \begin{split}
    \hat{S}_T(x) &= \mathop{\mathrm{argmax}}\limits_{k\in K}p(k_x|\bm{I}_x,\bm{\phi}) \\
    &= \mathop{\mathrm{argmax}}\limits_{k\in K}\left\{\pi_{kx}\prod\limits_{i=1}^{N_I}p(I_{i,x}|k_x,\phi_i)\right\}.
  \end{split}
  \label{eq:mas}
\end{equation}
In practice, the MAS result with $N_I\times t$ atlases can be generated from $t$ times of groupwise registration over $N_I$ subjects followed by label fusion using \cref{eq:mas}.

\section{Experiments and Results}\label{sec:experiments&results}
In this section, we investigate two applications of the proposed framework described in \cref{sec:applications}. In both of the two experiments, the neural networks were trained on a $\text{NVIDIA}^\circledR$ $\text{RTX}^\textnormal{TM}$ 2080 Ti GPU with the spatial transformer module adapted from open-source code in VoxelMorph~\cite{journal/tmi/Balakrishnan2019}, implemented in TensorFlow~\cite{arxiv/Abadi2015}.
The Adam optimizer was adopted~\cite{proceedings/ICLR/Kingma2014}, with a cyclical learning rate bouncing between 1e-5 and 1e-4 to accelerate convergence and avoid shallow local optima~\cite{proceedings/WACV/Smith2015}.

\subsection{pMvMM-RegNet for SAS on whole heart MRI}
\subsubsection{Materials and baseline.}
This experiment was performed on the MM-WHS challenge dataset, which provides 120 multi-modality whole-heart images from multiple sites, including 60 cardiac CT and 60 cardiac MRI~\cite{journal/media/Zhuang2016, journal/media/Zhuang2019}, of which 20 subjects from each of the modalities were selected as training data.
Intra- (MR-to-MR) and inter-modality (CT-to-MR) but inter-subject registration tasks were explored on this dataset, resulting in 800 propagated labels in total for 40 test MR subjects.

An optimal weighting of bending energy could lead to a low registration error, when maintaining the global smoothness of the deformations.
To be balanced, we set $\lambda=0.5$ as the default regularization strategy\footnote{See Fig. 1 in the supplementary material for an empirical result.}.
We analysed different variants of the appearance model described in \cref{sec:appearance}, i.e. "MOG", "Mask", "NCC" and "ECC", and compared with a reimplementation of~\cite{journal/media/Hu2018}, known as "WeaklyReg", which exploited the Dice similarity metric for weakly-supervised registration.
In addition, with the propagated labels obtained from pairwise registrations, we evaluated the performance of MAS by applying a simple majority vote to the results, denoted as "MVF-MvMM".

\subsubsection{Results and discussion.}
\begin{table}[tbp]
  \small
  \centering
  \caption{Average substruture Dice and Hausdorff distance (HD) of MR-to-MR and CT-to-MR inter-subject registration, with * indicating statistically significant improvement given by a Wilcoxon signed-rank test ($p<0.001$).}
  \begin{tabular}{p{2.5cm}|L{2.2cm}|L{2.2cm}|L{2.2cm}|L{2.2cm}}
    \toprule
    \multirow{2}*{\textbf{Methods}} & \multicolumn{2}{C{4cm}|}{\textit{MR-to-MR}} & \multicolumn{2}{C{4cm}}{\textit{CT-to-MR}} \\
    \cline{2-5}
    & \multicolumn{1}{c|}{Dice} & \multicolumn{1}{c|}{HD (mm)} & \multicolumn{1}{c|}{Dice} & \multicolumn{1}{c}{HD (mm)} \\
    \hline
    WeaklyReg & $0.834\pm 0.031$ & $17.45\pm 2.482$ & $0.842\pm 0.033$ & $17.99\pm 2.681$ \\
    \hdashline
    Baseline-MOG & $0.832\pm 0.027$ & $19.65\pm 2.792$ & $0.851\pm 0.028$* & $19.03\pm 2.564$ \\
    Baseline-Mask & $0.840\pm 0.028$* & $16.91\pm 2.374$* & $0.851\pm 0.032$ & $17.51\pm 2.687$* \\
    Baseline-ECC & $0.844\pm 0.026$* & $16.69\pm 2.355$* & $0.850\pm 0.032$ & $17.70\pm 2.659$ \\
    Baseline-NCC & $0.847\pm 0.028$* & $16.83\pm 2.422$ & $0.850\pm 0.032$ & $17.78\pm 2.721$ \\
    \hline
    MVF-MvMM & \multicolumn{2}{C{4cm}|}{Dice=$0.871\pm 0.025$} & \multicolumn{2}{C{4cm}}{HD (mm)=$17.21\pm 4.408$} \\
    \bottomrule
  \end{tabular}
  \label{tab:reg_dice}
\end{table}

\cref{tab:reg_dice} presents the Dice statistics of both intra- and inter-modality registration tasks on the MM-WHS dataset.
With increasingly plausible modelling of the relationship between appearance and anatomy, we have observed better registration accuracy especially for MR images, indicating efficacy of the proposed framework.
Fusing labels by majority vote ("MVF-MvMM") can produce a better segmentation accuracy, reaching an average Dice score of $0.871\pm  0.025$\footnote{See Fig.2 in the supplementary material for evaluation statistics on all cardiac substructure.}, comparable to the inter-observer variability of $0.878\pm 0.014$ reported in~\cite{journal/media/Zhuang2019}.

\subsection{gMvMM-RegNet for MAS on LGE CMR}
\subsubsection{Materials and baseline.}
In this experiment, we explored MAS with the application of \cref{eq:mas} on MS-CMRSeg challenge dataset~\cite{journal/pami/Zhuang2019}.
The dataset consists of 45 patients scanned using three CMR sequences, i.e. the LGE, T2 and bFFSP, from which 20 patients were chosen in random for training, 5 for validation and 20 for testing.
We implemented inter-subject and inter-modality groupwise registration and evaluated the MAS results on LGE CMR images.

A 2D version of the network architecture described in \cref{sec:network} was devised to jointly predict the deformation fields for $N_I$ atlases by optimizing \cref{eq:loss}.
The MAS result was generated by $t$ times of groupwise registration over $N_I$ randomly sampled subjects followed by label fusion using \cref{eq:mas}.

\subsubsection{Results and discussion.}
\begin{figure}[t]
  \centering
  \includegraphics[width=\textwidth]{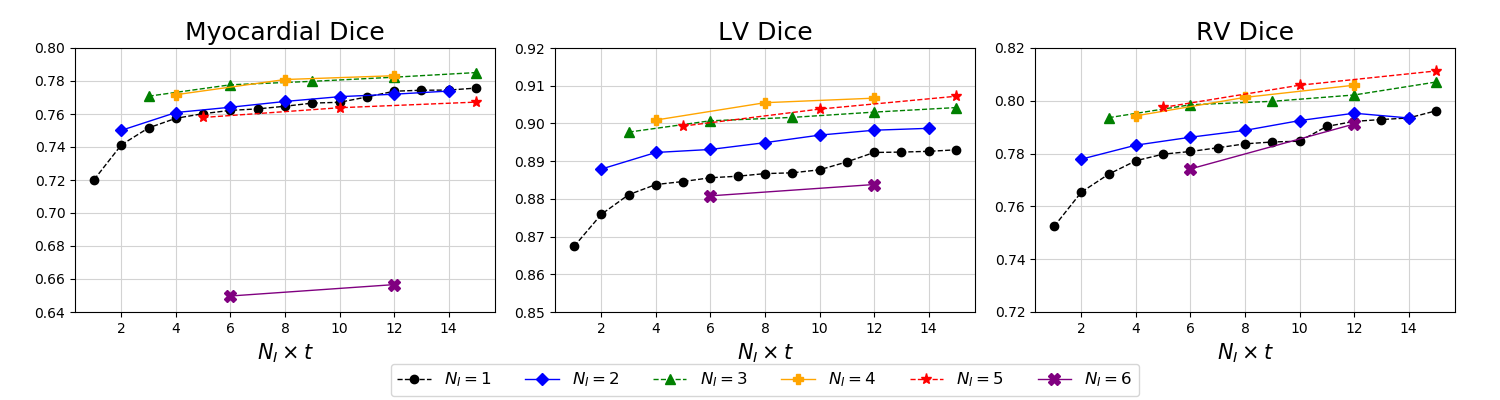}
  \caption{Dice scores of MAS results using $N_I\times t$ atlases, where $N_I$ denotes the number of subjects used in each groupwise registration and $t$ counts the number of groupwise registrations performed before label fusion.}
  \label{fig:mas_dice}
\end{figure}

The comparison between SAS and MAS highlights that more accurate and realistic segmentation is generated by groupwise registration than pairwise registration, especially for apical and basal slices\footnote{See Fig. 3 in the supplementary material for visualization of the segmentation results.}.
\cref{fig:mas_dice} further reports the mean Dice scores for each cardiac substructure obtained from MAS using $t$ times of groupwise registration with $N_I$ subjects.
With a fixed total number of atlases, label fusion on 2D slices resulting from groupwise registration outperforms those from conventional pairwise registration, reaching the average myocardium Dice score of $0.783\pm 0.082$.
However, we also observe decline in accuracy when having a large number of subjects ($N_I \geq 5$) to be groupwise registered.
This discrepancy could be attributed to the lack of network parameters compromising the predicted deformations.

\section{Conclusion}\label{sec:conclusion}
In this work, we propose a probabilistic image registration framework based on multivariate mixture model and neural network estimation, coupling groupwise registration and multi-atlas segmentation in a unified fashion.
We have evaluated two applications of the proposed model, i.e. SAS via pairwise registration and MAS unified by groupwise registration, on two publicly available cardiac image datasets and compared with state-of-the-art methods.
The proposed appearance model along with MvMM has shown its efficacy in realizing registration on cardiac medical images characterizing inferior intensity class correspondence.
Our method has also proved its superiority over conventional pairwise registration algorithms in terms of segmentation accuracy, highlighting the advantage of groupwise registration as a subroutine to MAS.

\section*{Acknowledgement}
This work was supported by the National Natural Science Foundation of China (grant no. 61971142).

\bibliographystyle{splncs04}
\bibliography{paper}

\end{document}